\documentclass[pdflatex,sn-mathphys-num]{sn-jnl}

\usepackage{graphicx}%
\usepackage{multirow}%
\usepackage{amsmath,amssymb,amsfonts}%
\usepackage{amsthm}%
\usepackage{mathrsfs}%
\usepackage[title]{appendix}%
\usepackage{xcolor}%
\usepackage{textcomp}%
\usepackage{manyfoot}%
\usepackage{booktabs}%
\usepackage{algorithm}%
\usepackage{algorithmicx}%
\usepackage{algpseudocode}%
\usepackage{listings}%

\theoremstyle{thmstyleone}%
\theoremstyle{thmstyletwo}%

\theoremstyle{thmstylethree}%

\begin{document}

\title[Evaluating passing decision-making in
professional football: An enhanced MPNN
approach to Receiver Selection]{Evaluating passing decision-making in
professional football: An enhanced MPNN
approach to Receiver Selection}


\author*[1]{\fnm{Gabriel} \sur{Masella}}\email{masellagabriel@gmail.com}

\author[2]{\fnm{Giuseppe Alessio} \sur{D'Inverno}}\email{gdinvern@sissa.it}

\author[3]{\fnm{Max} \sur{Goldsmith}}\email{max.goldsmith@rbfa.be}

\author[2]{\fnm{Gianluigi} \sur{Rozza}}\email{grozza@sissa.it}

\affil*[1]{\orgdiv{Department of Mathematics, Informatics and Geoscience}, \orgname{University of Trieste}, \orgaddress{\street{via Weiss 2}, \city{Trieste}, \postcode{34128}, \country{Italy}}}

\affil[2]{\orgdiv{MathLab}, \orgname{International School for Advanced Studies (SISSA)}, \orgaddress{\street{Via Bonomea 265}, \city{Trieste}, \postcode{34136}, \country{Italy}}}

\affil[3]{\orgname{Royal Belgium Football Association}, \orgaddress{\street{Rue de Bruxelles 480}, \city{Tubize}, \postcode{1480}, \country{Belgium}}}


\abstract{The process of decision-making in football is characterized by a complex interplay between spatial positioning, opponent pressure, and player intent. This work introduces a Graph Neural Network (GNN) framework designed to predict Receiver Selection, the optimal passing target, by modeling on-field interactions as dynamic graphs. Each player is represented as a node with positional and contextual features, while potential passing lines form weighted edges characterized by distance, angle, and pressure metrics. A Message-Passing Neural Network (MPNN) has been developed and trained using a combination of tracking data and event data from professional matches, synchronized through a robust pipeline based on an optimized version of the Needleman-Wunsch Algorithm. The model achieves competitive accuracy in identifying the actual chosen receiver and state-of-the-art accuracy within its top three suggestions. 
Our model further offers quantification of each option's likelihood, threat, and creativity, enabling performance analysts to evaluate over 1,000 passes in seconds.}


\maketitle

\section{Introduction}
\label{intro}

The application of data science techniques within the sports industry marks a paradigm shift from an intuitive to an evidence-based approach.
This transformation was originally pioneered in baseball through the concept of \textit{sabermetrics}, a term popularized by the Moneyball philosophy \cite{silva2016sports, lewis2003moneyball}.
In essence, this movement emphasized the potential of employing objective statistical analysis to enhance performance, leading to insights that traditional scouting methods had failed to capture.
Football, in its traditional form, initially resisted this shift due to its fluid, low-scoring, and continuous nature, but the modern era has been defined by a "big data" revolution.\\
\\
Quantitative analysis offers a comprehensive understanding of the sport of football; however, passing strategy is a crucial aspect that requires further attention \cite{power2017not,Anzer2022}.
This fundamental skill is the foundation for maintaining possession of the ball, progressing towards the goal, and creating scoring opportunities.
The act of passing is not limited to pure technical execution; it involves a dynamic and contested environment with multiple potential recipients in which players must rapidly evaluate spatial opportunities against immediate defensive pressure.
Despite its importance, the identification of an "optimal" passing decision remains a challenging problem in football analytics.
Traditional metrics, including completion rate and expected pass completion probability ($xPass$), offer a limited perspective on the quality of the decision itself \cite{Anzer2022}.
For instance, a successful lateral pass under minimal pressure is often prioritized based on basic statistics over an ambitious, unsuccessful forward pass into a defensive block, even if the second option represented a higher-value tactical choice.
Conventional approaches typically evaluate only the executed action, overlooking the alternative options available at the moment of release.
In order to accurately measure the process of decision-making, a model must consider not only the outcome of a decision but also the potential opportunity that was foregone by choosing an alternative option.\\
\\
To address this research gap, existing literature has moved towards more sophisticated spatial modeling.
Physics-based models \cite{spearman2017physics} have been employed to estimate pass probability based on interception times, while Expected Possession Value (EPV) frameworks have begun to assign value to specific field locations \cite{Anzer2022,fernandez2019decomposing}.
In contemporary research, Graph Neural Networks (GNNs) have emerged as a potent instrument for modeling the relational interactions between players \cite{bekkers2024graph}.
Many existing architectures, however, rely on fully connected graphs, which require a quadratic cost of computations, or ball-centric approaches, which frequently lead to "over-squashing" during the message-aggregation process, resulting in the loss of critical contextual information\cite{alon2021on,didoes}.\\
\\
In this paper, we propose the \emph{Football Pass MPNN} framework, a model that represents the football pitch as a dynamic graph to explicitly capture the geometric and tactical constraints of passing. By leveraging the inductive biases inherent in geometric deep learning, our approach transcends the limitations of static snapshots, enabling real-time analysis of the potential of every passing lane. The following contributions are made:
\begin{itemize}
    \item \textbf{Graph-Based Representation}: A "star graph" topology is introduced in which the passer assumes the role of the central node.
    This formulation explicitly encodes relational properties, such as obstruction cones and opponent pressure, directly into the edges and nodes of the graph.
    \item \textbf{Edge-Conditioned Message Passing}: The development of a Message Passing Neural Network (MPNN) that employs edge-conditioned reasoning to predict the optimal receiver has been implemented, achieving a Top-3 Accuracy of $97.80\%$ and improving over traditional baseline models.
    \item \textbf{Advances in Decision Metrics}: Interpretable metrics, such as \textit{"Best Pass Score"}, \textit{"Good Pass Score"} and \textit{"Creativity Ratio"}, are defined in order to translate probabilistic outputs into actionable insights for scouting and performance analysis.
\end{itemize}
The remainder of this paper is organized as follows: Section \ref{section:related_works} reviews related work; Section \ref{section:data_and_graph} details the dataset and graph formulation; Section \ref{section:football_MPNN} describes the model; Section \ref{section:results} presents experimental results; Section \ref{section:applications} discusses applications; and Section \ref{section:conclusions} concludes the paper.

\section{Related Works}
\label{section:related_works}
The existing literature on pass models, decision-making, and geometric deep learning in football is extensive and heterogeneous.\\
The development of these models follows a clear progression, beginning with the physics of pass completion, advancing to the evaluation of executed actions, and culminating in the prediction of, and reasoning behind, the passer's decision among multiple alternatives.\\
\\
Early research focused on the fundamental question of pass probability. Spearman et al. \cite{spearman2017physics} were among the first to develop physics-inspired models of time-to-intercept and time-to-control, allowing the estimation of pitch control surfaces and pass completion probabilities. While these models were capable of determining which player could reach a pass first with a reasonable degree of accuracy (approximately 68\%), they primarily evaluated physical availability rather than tactical intent. This theoretical foundation was later expanded by Dick et al. \cite{Dick2022}, who combined movement models with ball dynamics to quantify a player's probabilistic availability to receive a pass. However, the model still stopped short of simulating the passer's deliberate choice from the set of all available teammates.\\
\\
Building on this, a subsequent line of research emerged to evaluate the consequences of these choices by developing frameworks for the valuation of specific on-ball actions.\\
The Expected Possession Value (EPV) model, proposed by Fernandez et al. \cite{fernandez2020soccermap, fernandez2019decomposing}, employed convolutional networks to generate spatial maps of pass success probability and expected value. 
In a similar study, the VAEP framework by Decroos et al. \cite{decroos2020vaep} provided an objective method for evaluating actions based on their impact on scoring and conceding probabilities.
However, these value-based approaches generally evaluate the executed pass in isolation, often ignoring the alternative options that were rejected. This is a critical omission when assessing the quality of decision-making. \\
To address this gap, recent research has attempted to develop a model of the decision-making process itself. Li and Zhang \cite{Li2019} proposed a framework for receiver prediction that they categorized as a learning-to-rank problem. They developed a LambdaMART model to rank available teammates, but their tabular approach lacked the capacity to explicitly model the spatial relationships between players.\\
\\
Recent advances in geometric deep learning have demonstrated that Graph Neural Networks (GNNs) are the most natural architecture for representing the relational interactions of the pitch \cite{bekkers2024graph}.\\
Stöckl et al. \cite{stockl2021making} employed graph convolutional networks (GCNs) to analyze defensive performance through models such as xThreat and xReceiver. This study demonstrated that graph-based approaches can effectively estimate the likelihood of each player receiving the ball. Rahimian et al. \cite{rahimian2023pass, Rahimian2026tgn} further expanded this research by utilizing Temporal Graph Networks (TGNs) to quantify pass reception probabilities against defensive structures, effectively capturing dynamic interactions over time.\\
Despite these advancements, a research gap remains in fully capturing the multi-dimensional, context-dependent constraints of a single decision snapshot. While existing GCN and TGN approaches are powerful, they often prioritize temporal sequences or fully connected topologies, which is problematic in terms of specific spatial reasoning.\\
In order to address this limitations, the present study adapts the MPNN framework, originally introduced by Gilmer et al. \cite{gilmer2017neural}, to explicitly model "edge-conditioned" reasoning, where the quality of each passing option is determined not just by the receiver, but also by the geometric features of the passing lane itself.

\section{Data and Graph Formulation}
\label{section:data_and_graph}
The development of a robust receiver prediction model requires a dataset that is not only large in volume but is also semantically rich and spatio-temporally complete.
The following section provides a comprehensive overview of the construction of the dataset, the mathematical conversion of raw coordinates into a geometric graph, and the theoretical justification for the selected graph topology.\\
\subsection{Dataset Overview}
\begin{figure}
    \centering
    \includegraphics[width=0.95\linewidth]{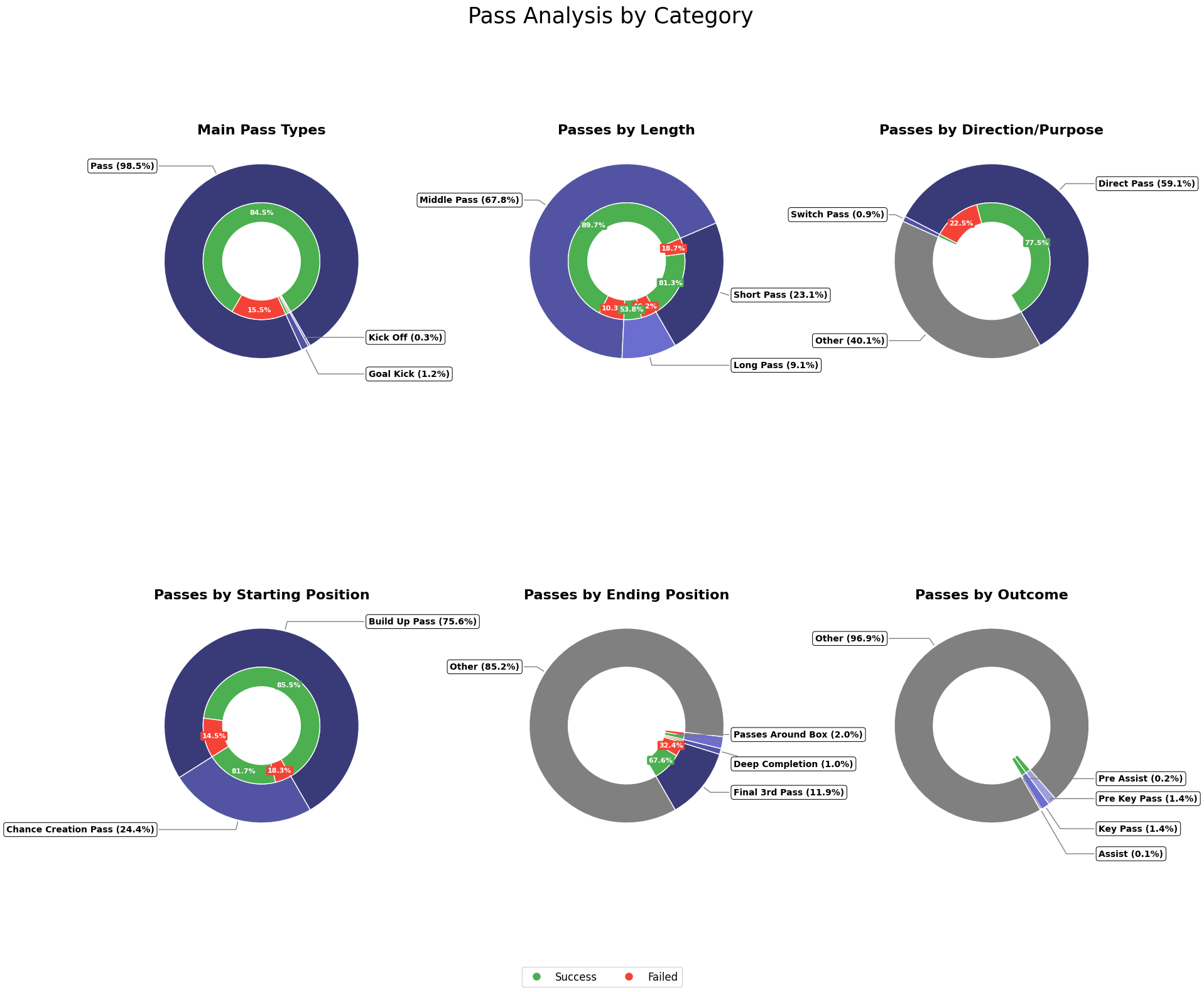}
    \caption{Analysis of the different categories of passes available in our dataset.}
    \label{fig:pass_categories}
\end{figure}
This research is based on a comprehensive dataset comprising 369 international matches from the \emph{UEFA Nations League (2024)} and \emph{FIFA World Cup Qualifiers (2026)}.\\
To ensure a comprehensive perspective of the game's state, two distinct data streams were combined: event data, which provides semantic labels such as passes and shots, and tracking data, which provides the continuous $(x, y)$ coordinates of all 22 players and the ball at a rate of 25 Hz.
The synchronization of these heterogeneous sources was achieved through the implementation of a dedicated pipeline, employing an optimized version of the Needleman-Wunsch algorithm\cite{needleman1970general}, proposed by Oonk et al. \cite{oonk2025sync}.
The synchronization process ensures that every discrete on-ball action is contextualized within the full spatio-temporal environment of the match. This approach allowed for the accurate mapping of discrete events to their corresponding tracking frames without the necessity for labeled training data, achieving a synchronization success rate of 91.3\% across the available match inventory.\\
The resulting dataset contains approximately 216,000 successful open-play passes that were utilized for training purposes. This synchronized collection of data, as shown in Figure \ref{fig:pass_categories}, demonstrates a remarkable level of semantic richness, encompassing a wide-ranging categorization of pass types that reflects the complex tactical landscape of modern football. 
\begin{itemize} 
    \item \textbf{Tactical Distribution}: The dataset is dominated by build-up passes (75.6\%) originating in the defensive and middle thirds, contrasted by high-leverage chance creation passes (24.4\%) in the final 40 meters.
    \item \textbf{Distance Dynamics}: Middle-distance passes (10–34m) constitute the core of possession (67.8\%) with a high success rate of 89.7\%, while long passes ($\ge34m$) represent tactical risks with a significantly lower completion rate of 53.8\%.
    \item \textbf{Progressive Intent}: Direct passes, designed to reduce the distance to the opponent's goal, account for 59.1\% of the data, illustrating a persistent drive for verticality despite a moderate success rate of 77.5\%.
    \item \textbf{Defensive Penetration}: High-value actions such as deep completions and passes around the box represent just 3.0\% of the total volume, reflecting the difficulty of operating in congested final-third spaces where success rates drop to nearly 50\%.
    \item \textbf{Elite Outcomes}: The dataset captures the rarest and most decisive moments in the game. These include key passes (1.4\%), which are defined as passes that result in a shot, and assists (0.1\%), passes that result in a goal, providing the necessary depth to model high-reward decision-making.
\end{itemize}
By integrating these diverse semantic layers ranging from routine lateral recycling to high-risk line-breaking balls, the dataset provides a detailed representation of the decision-making variables that players face. This granularity allows the Football Pass MPNN to learn not only the physics of a pass, but also the underlying tactical logic governing why certain options are preferred over others in specific match contexts.

\subsection{Mathematical Formulation}
The problem of receiver selection can be defined as the learning of a mapping function, denoted by $\Phi$, which approximates the conditional probability distribution of a passer selecting a specific teammate given the state of the game.\\
Let $\mathcal{S}$ represent the state of the game at the precise moment $t$ a pass is initiated. This state consists of the coordinates of the attacking team $\mathcal{A}$, the defending team $\mathcal{D}$, and the ball $b$. The attacking team is composed of a passer, denoted by $v_p$, and a set of potential receivers, denoted by $V_{rec} = \mathcal{A} \setminus \{v_p\}$.
The objective is to learn a parameterized function $F_\theta: \mathcal{S} \to \Delta^{|V_{rec}|-1}$, where $\Delta$ is the probability simplex, such that:
\begin{equation}
P(y = v_i | \mathcal{S}) \approx F_\theta(\mathcal{G}(\mathcal{S}))_i
\end{equation}
Where:
\begin{itemize}
    \item $y$ is the random variable representing the target receiver.
    \item $\mathcal{G}(\mathcal{S})$ is the graph transformation of the state.
    \item $\theta$ are the learnable parameters of the MPNN.
\end{itemize}
The ground truth dataset is defined as consisting exclusively of successful open-play passes. It is hypothesised that in elite professional football, a completed pass should be considered as a valid proxy for a "correct" or "optimal" decision. While this approach eliminates theoretically optimal passes that failed due to execution error, it provides a consistent, high-quality supervisory signal for learning tactical patterns, treating the player as an expert agent whose successful choices maximize an implicit reward function, such as retaining possession or gaining territory.\\
\\
\begin{figure}
    \centering
    \includegraphics[width=0.95\linewidth]{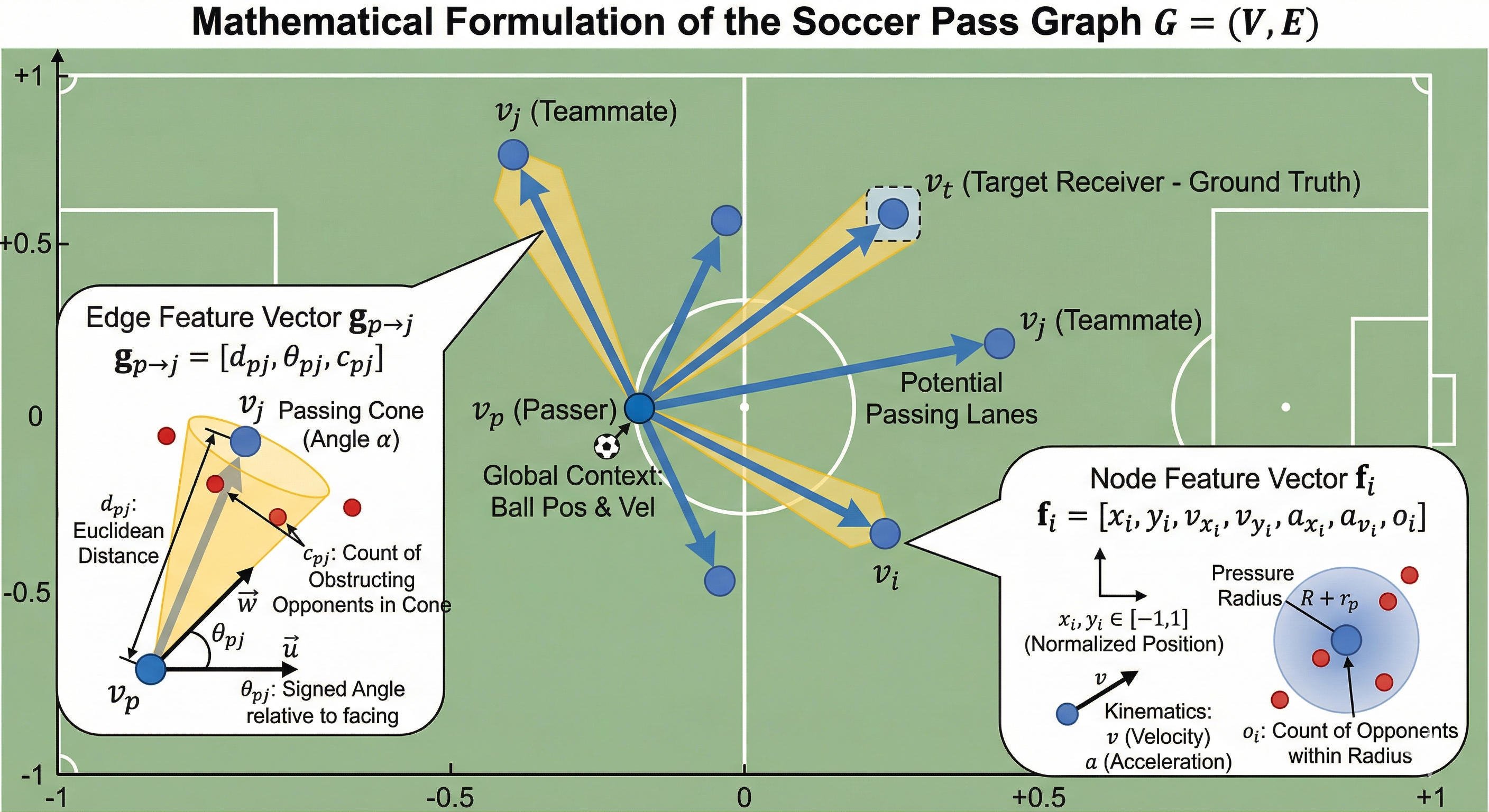}
    \caption{Illustration of the conversion from pass data to graph, with the mathematical background}
    \label{fig:pass_graph}
\end{figure}
In order to process the state $\mathcal{S}$ using Geometric Deep Learning, it is necessary to transform it into a directed graph $G = (V, E)$, where the topology is explicitly designed to model the passer's decision-making process.\\
The set of nodes $V$ corresponds to the 11 players of the attacking team. Each node $v_i \in V$ is associated with a feature vector $\mathbf{f}_i \in \mathbb{R}^7$ encoding its kinematic and tactical state:
\begin{equation}
\mathbf{f}_i = [x_i, y_i, v_{x_i}, v_{y_i}, a_{x_i}, a_{y_i}, c_i]
\end{equation}
where:
\begin{itemize}
    \item $x_i, y_i \in [-1, 1]$ are normalized Cartesian coordinates.
    \item $\vec{v}_i = (v_{x_i}, v_{y_i})$ and $\vec{a}_i = (a_{x_i}, a_{y_i})$ represent the velocity and acceleration vectors, capturing the player's momentum and movement intent.
    \item $c_i$ quantifies the \textit{Pressure} on that player, defined as the count of opponents within a local neighborhood of player $v_i$. Let $d(\cdot, \cdot)$ be the Euclidean distance and $R$ be a fixed distance threshold. Then $c_i = |\{o_k \in \mathcal{D} : d(v_i, o_k) \le R\}|$.
\end{itemize}
The set of edges $E$ represents the potential passing lanes. In our star graph topology, edges exist strictly from passer $v_p$ to each of their teammates $v_j$, denoted as $e_{p\to j}$. The defensive structure, which other models represent as separate nodes, is encoded directly into the edge attributes of the attacking graph. This reduces computational complexity while explicitly defining the difficulty of the pass.
Each edge is defined by a vector $\mathbf{g}_{p \to j} \in \mathbb{R}^3$:
\begin{equation}
\mathbf{g}_{p\to j} = [d_{pj}, \theta_{pj}, c_{pj}]
\end{equation}
Where:
\begin{itemize}
    \item $d_{pj} = \|\vec{p}_j - \vec{p}_p\|_2$ is the Euclidean distance.
        \item $\theta_{pj}$  is the signed angle between the passer's facing direction and the vector to the teammate, capturing the difficulty of body re-orientation. Let $\vec{u}$ be the normalized vector of the passer's facing direction (from ball to passer), and $\vec{w} = \vec{p}_{v_j} - \vec{p}_{v_p}$ be the vector from passer to teammate. The angle is computed as:
    \begin{equation}
        \theta_{pj} = \text{atan2}(\vec{u}_x \vec{w}_y - \vec{u}_y \vec{w}_x, \vec{u} \cdot \vec{w})
    \end{equation}
    \item $c_{pj}$ mathematically quantifies the "traffic" along a passing lane. The passing lane is modeled as a geometric cone with apex at $\vec{p}_{v_p}$, direction vector $\vec{w}$, and angular width $\alpha$. An opponent $o_k \in \mathcal{D}$ at position $\vec{p}_{o_k}$ is considered an obstruction if their physical presence intersects this cone. Let $r_p$ denote the effective player occlusion radius. Formally, let $\vec{t} = \vec{p}_{o_k} - \vec{p}_{v_p}$ be the vector from the passer to the opponent. The opponent is counted if and only if:
    $$\arccos\left(\frac{\vec{w} \cdot \vec{t}}{\lVert \vec{w} \rVert \lVert \vec{t} \rVert}\right) - \arcsin\left(\frac{r_p}{\lVert \vec{t} \rVert}\right) \leq \frac{\alpha}{2} \quad \land \quad \lVert \vec{t} \rVert \leq \lVert \vec{w} \rVert + r_p$$
    The first condition ensures angular intersection, while the second ensures the opponent is located between the passer and the receiver (finite length).
\end{itemize}
This rigorous formulation ensures that the MPNN does not merely learn from the positions of teammates, but conditions its message passing on the explicit geometric viability of the passing lane itself.

\subsection{Topological Choices and Comparisons}
Further consideration should be focused on the construction of the graph, which encodes distinct inductive biases and practical tradeoffs for modeling passing decisions.\\
A prevailing approach, adopted by Stöckl et al. \cite{stockl2021making} and Rahimian et al. \cite{rahimian2023pass, Rahimian2026tgn} in this field, involves the construction of 22-node (23-node graphs, including the ball) fully connected graphs with team flags.
In terms of its advantages, the method is highly expressive, with the ability to explicitly capture every pairwise relation. However, this approach comes with a quadratic $O(N^2)$ computational cost and an increased risk of overfitting. \\
Another pattern that was observed in the study conducted by Bekker et al. \cite{bekkers2024graph} was the use of a ball-centered two-layer design, in which the players are connected to the ball's nodal point and to their teammates, facilitating the consolidation of information through a centralized hub. The consequence of this choice is that the system could lead to a phenomenon known as over-squashing of messages \cite{alon2021on, didoes}.
This term refers to the compression of contextual signals from distant nodes as they pass through the central node, leading to a reduction in the effective flow of information.\\ 
The design implemented in this study led to the formulation of a star graph, as illustrated in Figure \ref{fig:pass_graph} and elucidated in Section \ref{section:data_and_graph}, in which the passer is positioned at the center and the ten teammates are connected as satellite nodes. Opponents and contextual elements are represented as hard-coded node and edge features, including relative position, velocity, pressure flags, and opponent-in-line.
This approach results in a compact adjacency matrix, which is advantageous due to its linear number of edges, enhanced computational efficiency, and a clear passer‑centric inductive bias that directly models the decision hub.\\
However, this approach is subject to certain limitations. In its practical implementation, the star graph adopts a balanced position by emphasizing passer-centric reasoning and demonstrating scalability; this can lead to an inaccurate representation of interactions between teammates.
To address this, a strategy of employing multiple message-passing rounds or utilizing edge-conditioned message functions was employed in this research. This approach preserves rich spatial and opponent cues, ensuring the accuracy of the system despite the sparse topology.

\section{Football Pass MPNN}
\label{section:football_MPNN}

The \textit{Football Pass MPNN} framework is implemented as a deep Message Passing Neural Network \cite{gilmer2017neural} designed to model the passer-centric decision hub of a football player. The architecture transforms the passing graph through three primary stages: feature embedding, iterative message passing, and readout.

\begin{figure}[ht]
    \centering
    \includegraphics[width=0.95\linewidth]{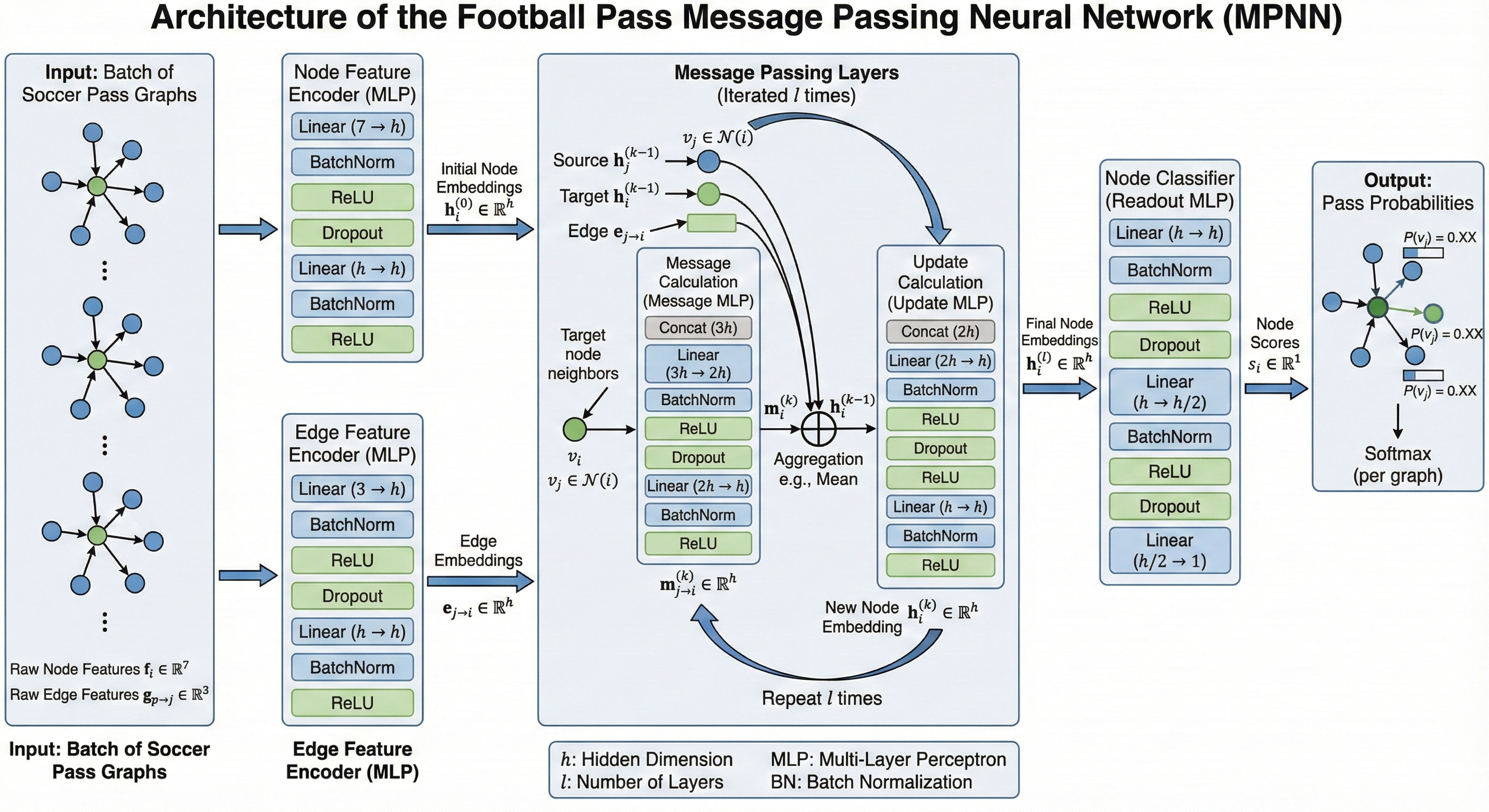}
    \caption{The complete architecture of the Football Pass MPNN.}
    \label{fig:model_architecture}
\end{figure}

\subsection{Architecture Overview}
To enable the network to learn complex non-linear representations from these physical quantities, the raw features matrix $\mathbf{V} \in \mathbb{R}^{N \times 7}$ and $\mathbf{E} \in \mathbb{R}^{M \times 3}$ are first projected into a shared latent dimension $h$ (embedding size) using distinct Multi-Layer Perceptrons (MLPs), denoted as $\phi_{\text{node}}$ and $\phi_{\text{edge}}$.
$$\mathbf{h}_i^{(0)} = \phi_{\text{node}}(\mathbf{f}_i), \quad \forall v_i \in V$$
$$\mathbf{e}_{ij}' = \phi_{\text{edge}}(\mathbf{g}_{ij}), \quad \forall (v_i, v_j) \in E$$
The core reasoning occurs during $L$ synchronous message-passing steps. In each layer $l$, an edge-conditioned message function $M_l$ computes a vector $\mathbf{m}_{j \to i}^{(l)}$ by considering the interaction between the passer, the receiver, and the passing lane difficulty:
\begin{equation}
    \mathbf{m}_{j \to i}^{(l)} = M_l \left( \mathbf{h}_j^{(l-1)} \parallel \mathbf{h}_i^{(l-1)} \parallel \mathbf{e}'_{ij} \right)
\end{equation}
These messages are aggregated via a permutation-invariant operator to update the node states:
\begin{equation}
    \mathbf{h}_i^{(l)} = U_l \left( \mathbf{h}_i^{(l-1)} \parallel \bigoplus_{j \in \mathcal{N}(i)} \mathbf{m}_{j \to i}^{(l)} \right)
\end{equation}
This iterative refinement allows the model to contextually evaluate each player's suitability relative to the defensive topology and the passer's momentum.

\subsection{Readout and Probability Estimation}
After $L$ message passing steps, the final node embeddings $\mathbf{h}_i^{(L)}$ contain a rich, context-aware representation of each player's suitability as a pass target.\\
The final phase is the readout, which maps these high-dimensional embeddings to a scalar score $s_i$ representing the unnormalized likelihood of receiving the pass.
$$s_i = R(\mathbf{h}_i^{(L)})$$
To accommodate the variable cardinality of team sports, a per-graph softmax is applied over the set of nodes within the specific graph to yield the final selection probability for each teammate $v_i \in V$:
\begin{equation}
    P(y = v_i | G) = \frac{\exp(s_i)}{\sum_{v_k \in V} \exp(s_k)}
\end{equation}
This formulation inherently handles the physical constraints of the sport, normalizing the scores across the available teammates regardless of numerical disparities, such as those caused by expulsions. The model is trained end-to-end to minimize the Cross-Entropy loss between the predicted distribution and the ground-truth receiver provided by the event data.

\subsection{Training and Hyperparameter Optimization}
The implementation and training of the model were executed using the PyTorch Geometric library \cite{Fey/etal/2025}, scaling efficiently to dynamic batching of variable-sized graphs. Optimization was performed via the AdamW algorithm, combined with a \texttt{ReduceLROnPlateau} learning rate scheduler to stabilize convergence.\\
To ensure architectural robustness, a Bayesian hyperparameter search was conducted utilizing the Optuna framework \cite{optuna_2019}. The configuration was systematically calibrated across the search space defined in Table \ref{tab:hyperparams}.\\
\begin{table}[ht]
    \centering
    \caption{Hyperparameter search space and best values found via Optuna.}
    \label{tab:hyperparams}
    \begin{tabular}{lcc}
        \toprule
        \textbf{Parameter} & \textbf{Search Range} & \textbf{Best Value} \\
        \midrule
        Hidden Dimension & $\{128, 256, 512\}$ & \textbf{512} \\
        Number of Layers & $\{3, 5, 7\}$ & \textbf{5} \\
        Dropout & $[0.1, 0.5]$ & \textbf{0.15} \\
        Learning Rate & $[10^{-5}, 10^{-3}]$ & \textbf{7.6e-4} \\
        Weight Decay & $[10^{-5}, 10^{-3}]$ & \textbf{5.7e-4} \\
        Batch Size & $\{64, 128, 256, 512\}$ & \textbf{64} \\
        Aggregation & $\{\text{mean}, \text{max}, \text{add}\}$ & \textbf{max} \\
        \bottomrule
    \end{tabular}
\end{table}
In terms of representational capacity, the hidden dimension exhibited the strongest positive correlation with validation accuracy ($+0.62$). The $512$-dimensional configuration consistently outperformed smaller architectures, suggesting that tactical passing decisions require substantial representational capacity to capture complex spatial interactions adequately. The correlation between the number of message-passing layers and overall performance was found to be relatively modest ($+0.15$). The optimal configuration utilized five synchronous layers. This indicates that within the compact star-graph topology, information propagates efficiently; additional layers increase the risk of over-smoothing node representations without yielding commensurate performance gains.\\
Regarding training dynamics, the learning rate was highly correlated with convergence ($+0.75$), with $7.6 \times 10^{-4}$ emerging as the optimal value. Furthermore, a strong inverse relationship with batch size was observed ($-0.59$). Smaller mini-batches of $64$ graphs facilitated more frequent parameter updates, enabling a more robust adaptation to the heterogeneous spatial configurations characteristic of the dataset. Regularization proved highly impactful on generalization capabilities: dropout showed a strong negative correlation ($-0.67$), indicating that excessive rates ($\ge 0.3$) substantially degraded performance by disrupting information flow across the narrow message-passing channels. Conversely, $L_2$ regularization via weight decay exhibited a moderate positive correlation ($+0.42$), demonstrating its efficacy in preventing the network from memorizing irrelevant positional correlations. Finally, the ``max'' message aggregator strictly outperformed both mean and summation operations. This outcome suggests that the model's efficacy is maximized when it isolates the most salient tactical features from its neighborhood, analogous to the cognitive filtering applied by elite players in time-constrained scenarios.

\section{Experimental Results}
\label{section:results}
\subsection{Performance Benchmarking}
To rigorously quantify the efficiency of the proposed approach, the available dataset was explicitly partitioned into training ($N=151,460$), validation ($N=32,456$), and unobserved test ($N=32,456$) sets, using a random split with a 70/15/15  ratio from the total size of the available passes. The Football Pass MPNN was evaluated on the test set against a suite of internal baseline models trained on the same dataset. The baselines include:\\
\begin{itemize}
    \item \textbf{Nearest Player Heuristic}: A foundational heuristic predicting the teammate closest to the passer in terms of Euclidean distance.
    \item \textbf{Logistic Regression}: A linear model trained on the full suite of node and edge features utilized by the MPNN.
    \item \textbf{LambdaMART \cite{Li2019}}: A tree-based learning-to-rank baseline replicating the specific feature subset described in the original LambdaMART paper (e.g., relative distances, angles, and goal coordinates), omitting the advanced relational components introduced in our architecture.
\end{itemize}
\begin{table}[h!]
    \centering
    \caption{Performance benchmarking on the unobserved test set ($N=32,456$)}
    \label{tab:baselines}
    \begin{tabular}{lcccc}
        \toprule
        \textbf{Model} & \textbf{Top-1 Acc. $\uparrow$} & \textbf{Top-3 Acc. $\uparrow$} & \textbf{AUROC $\uparrow$} & \textbf{Brier  $\downarrow$} \\
        \midrule
        Nearest Player Heuristic & 27.58\% & 63.85\% & 0.763 & 0.086 \\
        Logistic Regression & 43.63\% & 79.67\% & 0.853 & 0.084 \\
        LambdaMART \cite{Li2019} & 44.62\% & 80.61\% & 0.863 & 0.069 \\
        Football Pass MPNN (ours) & \textbf{75.83\%} & \textbf{97.80\%} & \textbf{0.963} & \textbf{0.036} \\
        \bottomrule
    \end{tabular}
\end{table}
The comparative analysis, detailed in Table \ref{tab:baselines}, demonstrates a substantial performance improvement. The linear regression model and the restricted tree-based approach (LambdaMART) achieved comparable Top-1 accuracies of approximately $44\%$, underscoring the limitations of modeling individual receiver suitability without accounting for complex geometric occlusion. In contrast, the MPNN achieves a robust Top-1 accuracy of $75.83\%$ and an overwhelming Top-3 accuracy of $97.80\%$, showcasing the inherent filtering capacity of the message-passing framework. Furthermore, the MPNN achieves a dominant global AUROC of $0.963$ and a Brier score of $0.036$, indicating a very strong capacity to distinguish the chosen receiver from non-targets.\\
\begin{table}[h!]
    \centering
    \caption{Performance comparison of different models from the literature}
    \label{tab:results_comparison}
    \begin{tabular}{lcccc}
        \toprule
        \textbf{xReceiver Models} & \textbf{Top-1 Acc. $\uparrow$} & \textbf{Top-3 Acc. $\uparrow$} & \textbf{AUROC $\uparrow$} & \textbf{Brier  $\downarrow$} \\
        \midrule
        Physics-Based \cite{spearman2017physics} & 0.68  & -- & 0.85 & -- \\
        LambdaMART \cite{Li2019} & 0.50 & 0.84 & -- & -- \\
        MLP Baseline \cite{stockl2021making} & 0.73 & -- & -- & -- \\
        GCN \cite{stockl2021making} & \textbf{0.83} & -- & -- & -- \\
        TGN \cite{rahimian2023pass} & 0.81 & -- & 0.85 & \textbf{0.010} \\
        Football Pass MPNN (ours) & 0.76 & \textbf{0.98} & \textbf{0.96} & 0.036 \\
        \bottomrule
    \end{tabular}
\end{table}
\\
In order to contextualize these capabilities, it is essential to take into consideration well-known architectures that have been evaluated in the literature on different datasets, as shown in Table \ref{tab:results_comparison}.\\
Traditional physics-based heuristics \cite{spearman2017physics} report Top-1 accuracies approaching $0.68$ and an AUROC of $0.85$. The foundational standalone MLP baseline outlined by St\"{o}ckl et al. \cite{stockl2021making} achieved a Top-1 of $0.73$. The proposed MPNN surpasses both, empirically validating the inclusion of explicit pitch relational structures.\\
Rahimian et al. \cite{rahimian2023pass} evaluated a Temporal Graph Network (TGN) that achieved a Top-1 accuracy of $0.81$, a ROC-AUC of $0.85$, and a Brier score of $0.010$. While the TGN outperformed our model in Top-1 accuracy and Brier score, our proposed MPNN achieves a significantly higher AUROC. Additionally, it is important to notice that the TGN framework \cite{Rahimian2026tgn} was evaluated on a narrow selection of specific scenarios, focusing exclusively on forward passes originating in the middle third of the pitch against structured defensive shapes, rather than a generalized open-play setting encompassing all directional pass variants.\\
In addition, although the Graph Convolutional Network (GCN) evaluated by St\"{o}ckl et al. \cite{stockl2021making} achieved a nominal accuracy of $0.83$, the highest in the known literature, this was achieved by implementing a weighted loss function (assigning a weight of 0.1 for non-targets and 1.0 for the true receiver), distorting the optimization landscape relative to standardized cross-entropy benchmarks.

\subsection{Inference Speed Analysis}

In addition to the evaluation of accuracy metrics, a foundational claim of the \textit{Football Pass MPNN} is that both the graph formulation and the model architecture were intentionally designed to be lightweight and highly optimized for real-time applications. In addition to the evaluation of accuracy metrics, a fundamental claim of the Football Pass MPNN is that both the graph formulation and the model architecture were intentionally designed to be lightweight and highly optimized for real-time applications. From the previously cited literature regarding spatial passing models, little attention has been given to the direct evaluation of inference speed, with the notable exception of the graph convolutional approach presented by St\"{o}ckl et al.\cite{stockl2021making}.This study, therefore, offers a useful reference point for the evaluation of real-time viability.\\
In order to conduct a fair comparison, a similar benchmarking experiment was designed. The evaluation simulated a sequential, real-time context by processing 1,000 randomly sampled, fully synchronized passes on a standard local notebook.\\
\begin{table}[h!]
    \centering
    \caption{Inference speed comparison expressed in seconds ($s$) as mean $\pm$ standard deviation.}
    \label{tab:inference_speed}
    \begin{tabular}{lccc}
        \toprule
        \textbf{Model Pipeline} & \textbf{Feature Crafting (s)} & \textbf{Inference (s)} & \textbf{Total Time (s)} \\
        \midrule
        St\"{o}ckl et al. \cite{stockl2021making} & $0.0440 \pm 0.0040$ & $\mathbf{0.0010 \pm 0.0002}$ & $0.0450 \pm 0.0060$ \\
        Football Pass MPNN (ours) & $\mathbf{0.0008 \pm 0.0001}$ & $0.0107 \pm 0.0017$ & $\mathbf{0.0115 \pm 0.0018}$ \\
        \bottomrule
    \end{tabular}
\end{table}
\\
In their analysis, the authors reported an average feature crafting time of $0.044 \pm 0.004s$ and an inference time of $0.001 \pm 0.0002s$, resulting in a total processing time of $0.045 \pm 0.006s$ per frame. While their pipeline is nearly capable of operating on a standard 25 Hz tracking data stream (which inherently requires processing within $0.040s$ per frame), this narrow mathematical margin provides insufficient allowance for the inevitable latency introduced by operational integration, such as API requests, network transmission, and data serialization.\\
On the other hand, the comparative results, presented in Table \ref{tab:inference_speed}, demonstrate a highly significant acceleration in data processing for our approach. By leveraging the analytically optimized star graph topology, the custom feature engineering process averaged merely $0.0008 \pm 0.0001s$. Thus, while the proposed 5-layer MPNN with a 512-dimensional hidden space requires a marginally longer inference time due to its expanded representational depth, the total pipeline execution is condensed to just $0.0115 \pm 0.0018s$ per frame on average. \\
Although absolute timings are subject to slight variations based on the specific hardware configurations utilized across different studies, the magnitude of this difference conclusively demonstrates the framework's efficiency. These results empirically confirm that the proposed methodology is substantially faster overall, possessing the computational elasticity to seamlessly process 25 Hz tracking streams in a live operational environment without vulnerability to serialization overheads.

\subsection{Tactical Key Performance Indicators}
\begin{figure}[h!]
    \centering
    \includegraphics[width=1\linewidth]{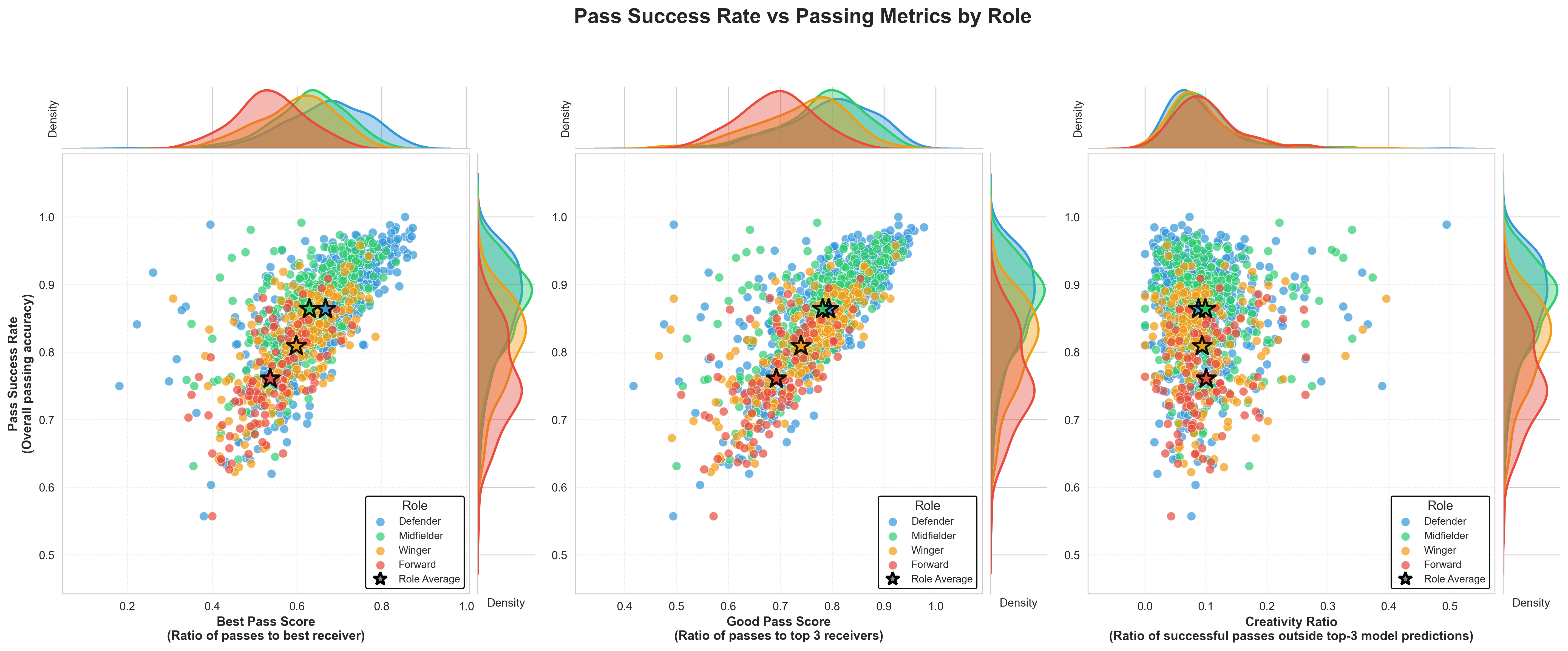}
    \caption{Kernel density estimations illustrating the risk-reward tradeoff relative to pass completion rates.}
    \label{fig:joint_plots}
\end{figure}

\begin{figure}[h!]
    \centering
    \includegraphics[width=1\linewidth]{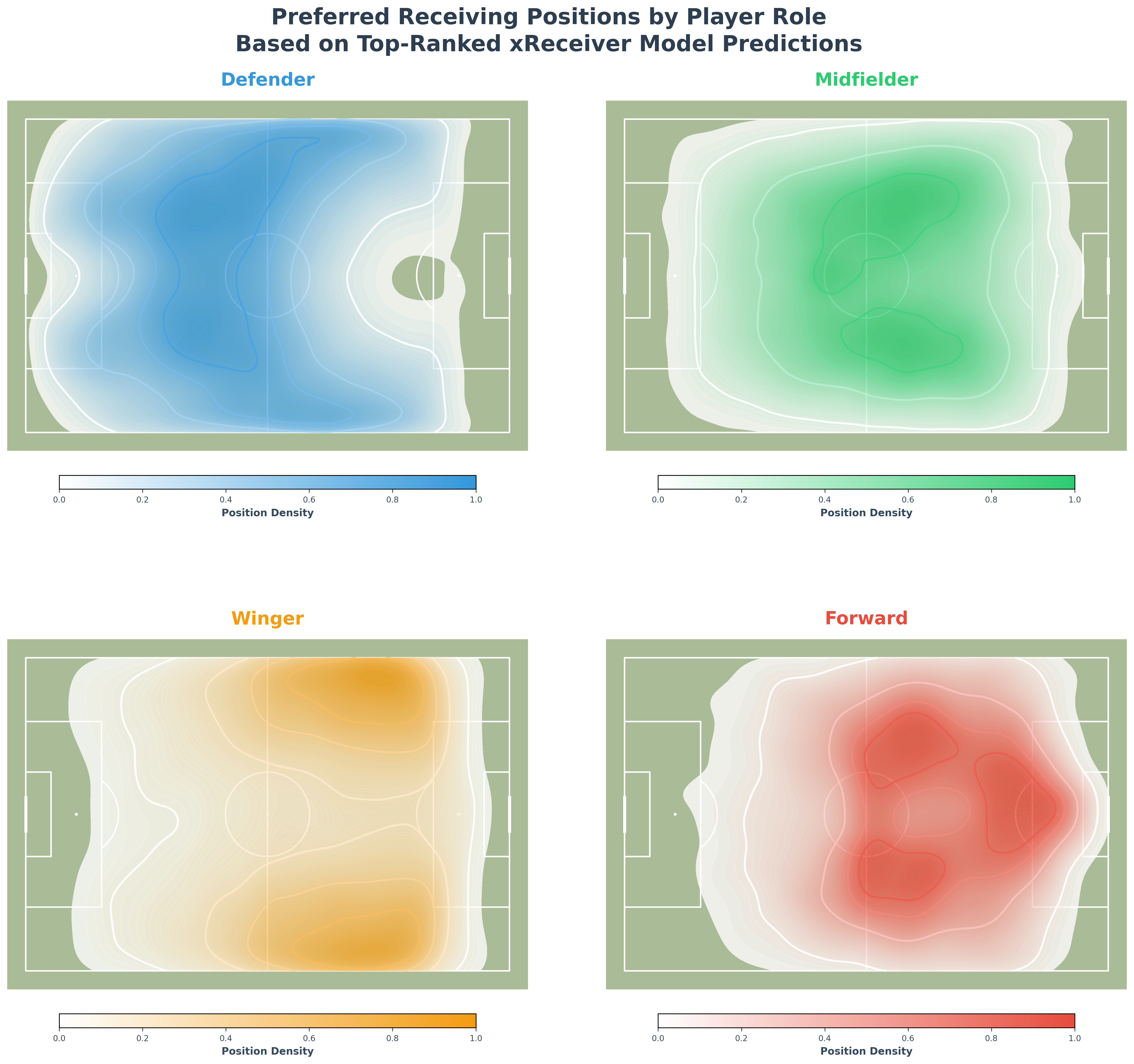}
    \caption{Density map of optimal receiver locations predicted by the model, categorized by positional role.}
    \label{fig:spatial_heatmaps}
\end{figure}

While high-ranking metrics confirm the model's statistical validity, the core objective is to translate abstract probability distributions into actionable coaching insights. To this end, the raw likelihood scores generated by the MPNN were projected into three interpretative Key Performance Indicators (KPIs):\\
\begin{itemize}
    \item \textbf{Best Pass Score (Reliability)}: The frequency with which a player successfully selects the model's Top-1 predicted target. This defines adherence to the mathematically optimal, lowest-risk option.
    \item \textbf{Good Pass Score (Safety)}: The rate at which the chosen receiver falls within the Top-3 predictions, acknowledging tactical soundness in restricted environments.
    \item \textbf{Creativity Ratio (Vision)}: The percentage of \textit{successful} passes directed to targets outside the model's top three predictions, isolating high-risk play-making that purposefully breaks mathematical expectation.
\end{itemize}
\begin{figure}[h!]
    \centering
    \includegraphics[width=1\linewidth]{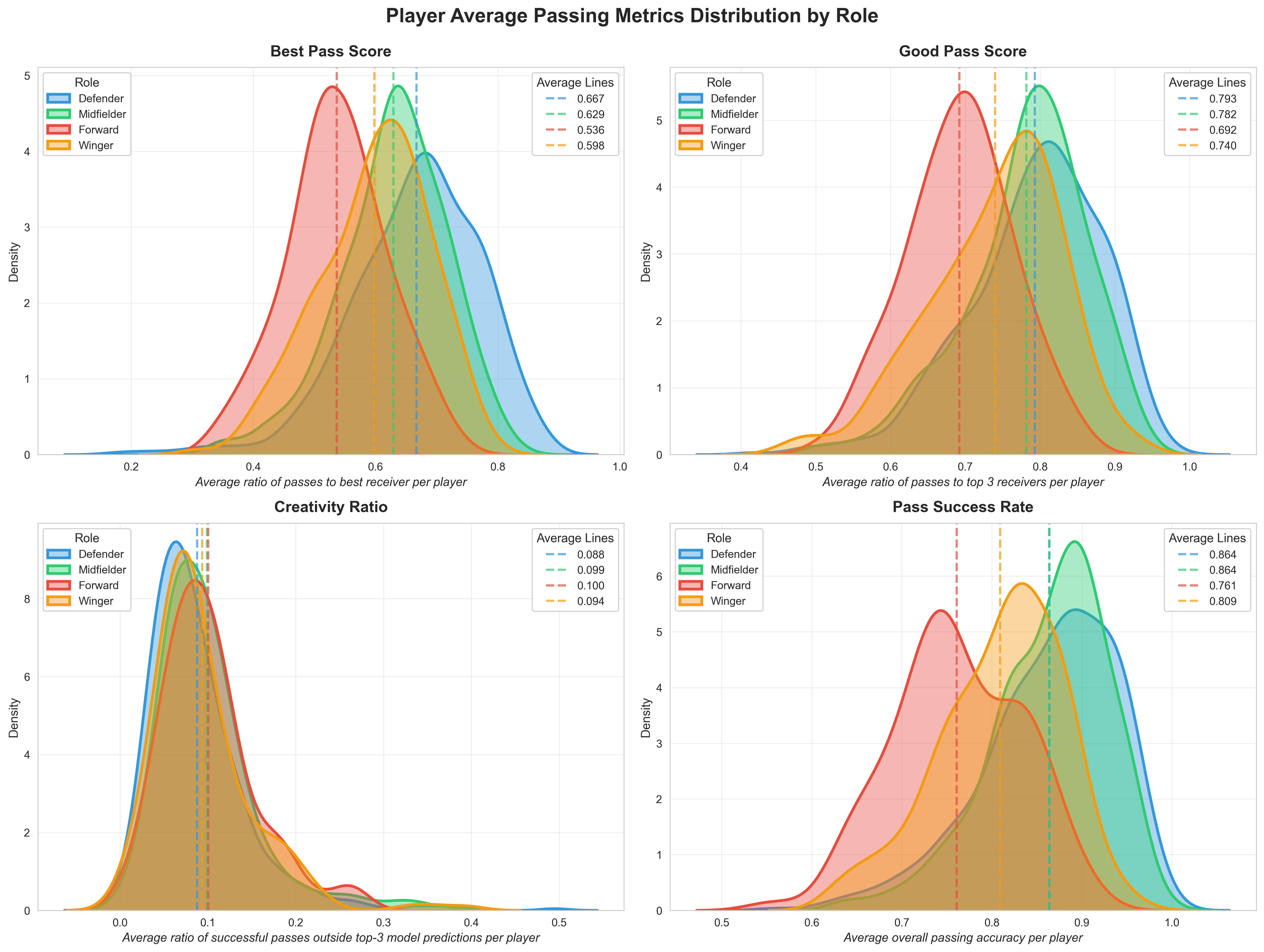}
    \caption{Metric distributions across operational roles based on players with at least 50 passes registered in the dataset.}
    \label{fig:kde_metrics}
\end{figure}
As illustrated in Figure \ref{fig:kde_metrics}, the KPI distributions reveal distinct patterns across playing roles. Defenders display the highest average Best Pass Score ($\mu \approx 0.667$), consistent with their primary task of safe possession circulation. Forwards show the lowest adherence ($\mu \approx 0.536$), as high-pressure zones frequently demand improvised selections that deviate from the model's optimal prediction. Wingers exhibit the most distinctive profile: despite recording high Creativity Ratios, they maintain elevated pass completion rates, suggesting a capacity for effective risk-taking that is qualitatively different from the lower-success creativity observed in forwards.\\
To validate these metrics against ground-truth outcomes, Figure \ref{fig:joint_plots} correlates each KPI with the player's overall pass success rate.
A strong positive correlation is confirmed between Best Pass Score and pass completion, while high Creativity Ratios are associated with completion rates in the $70$–$80\%$ range, quantitatively capturing the inherent risk-reward trade-off of unpredictable play-making.\\
Finally, to assess whether the model has internalized the spatial geometry of the game, Figure \ref{fig:spatial_heatmaps} visualizes the aggregate locations where each positional role is most frequently predicted as the optimal receiver.
Without any explicit positional constraints, the model reconstructs canonical receiving zones: defenders in the defensive half and wide channels, midfielders across the central box-to-box corridor, wingers along the attacking touchlines, and forwards centrally in front of the penalty area. This emergent spatial coherence confirms that the MPNN has genuinely learned the tactical geometry of the game from relational data alone.

\section{Applications}
\label{section:applications}

Beyond its theoretical contributions, the \textit{Football Pass MPNN} was designed to serve as an operational tool within the Data Department of the Royal Belgian Football Association (RBFA). The model's outputs are operationalized through a two-stage reporting pipeline: a \textit{Post-Game Review} that flags suboptimal passing decisions for individual player feedback, and a \textit{Pre-Match Scouting} module that profiles opposing players using the KPIs defined in Section \ref{section:results}. The computational efficiency of the architecture ensures both stages can be executed rapidly on incoming match data.

\subsection{Evaluating Decision-Making}
A primary application of the model resides in the objective evaluation of passing decisions. By comparing the actual choice made by a player against the predicted probability distribution, analysts are equipped to identify instances of suboptimal decision-making.\\
\begin{figure}[ht]
    \centering
    \includegraphics[width=0.95\linewidth]{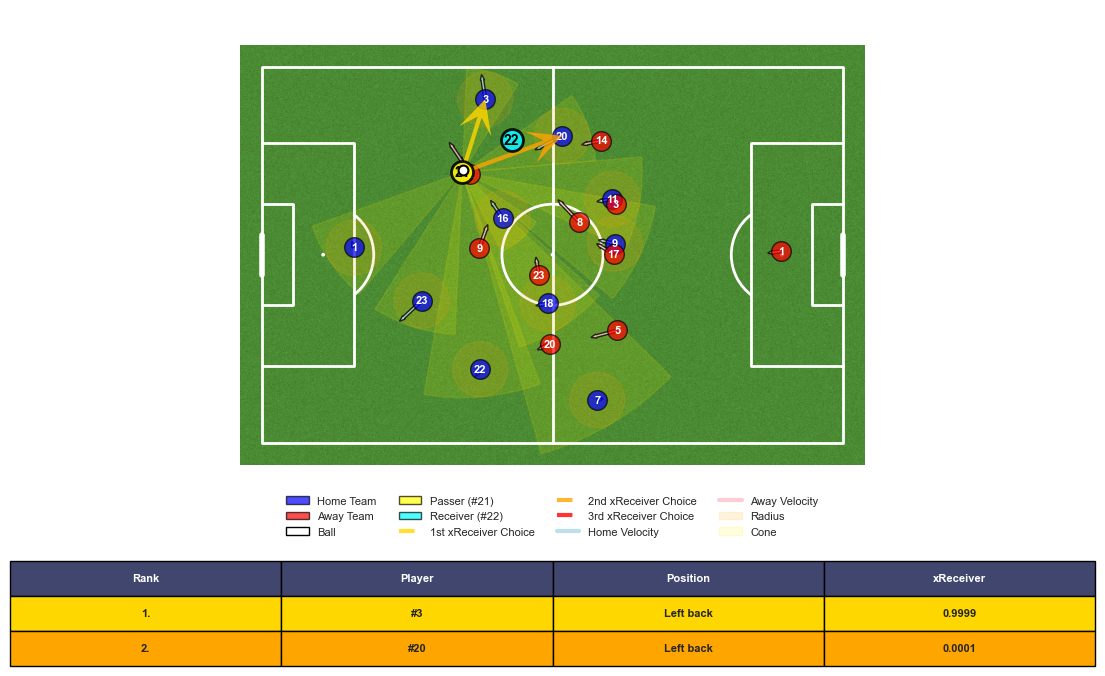}
    \caption{Visual output of the Football Pass MPNN analyzing an unsuccessful pass.}
    \label{fig:mpnn_fail}
\end{figure}
\\
As illustrated in Figure \ref{fig:mpnn_fail}, a concrete example of this application is presented. In the scenario under consideration, the passer (Node 21, colored in yellow) attempts a pass to a teammate (Node 20), which is subsequently intercepted by an opponent (Node 22, colored in cyan), resulting in a loss of possession. The MPNN offers a critical perspective on this sequence. As indicated in the accompanying prediction table, the model assigned a near-certain probability ($>0.99$) to a different teammate (Node 3, indicated by the yellow arrow) as the optimal target. This player was positioned advantageously to advance the play with minimal defensive pressure. Conversely, the executed choice was identified as a significantly suboptimal alternative.\\
Furthermore, the model's restricted probability distribution demonstrates its capacity for contextual understanding. Kinematic factors implicitly filter out tactically illogical or technically unfeasible targets, allowing performance analysts to focus exclusively on the disparity between the optimal computational solution and the human execution.

\subsection{Operational Integration}
To ensure the practical utility of these metrics, a dedicated pipeline was implemented to process incoming match data and generate structured reports for the coaching staff.

\subsubsection{Post-Game Review: Error Analysis}
Following the conclusion of a match, the system automatically identifies passes that could have been optimized. These are formally defined as unsuccessful passes where the MPNN identified an alternative option with a significantly higher probability of success. These instances are compiled into a Post-Game Review Report, where each entry is accompanied by its unique event identifier, video frame reference, and a 2D spatial visualization (analogous to Figure \ref{fig:mpnn_fail}). This automated workflow substantially reduces the manual effort required by video analysts to locate coaching moments, providing a curated list of decision-making errors suitable for individualized player feedback.

\subsubsection{Pre-Match Opponent Analysis}
In preparation for upcoming fixtures, the model is employed to evaluate the historical performance profiles of opposing players. By adopting the intuitive Key Performance Indicators established in Section \ref{section:results} (Best Pass Score, Good Pass Score, and Creativity Ratio), complex probabilistic outputs are made accessible to tactical staff. By identifying statistical outliers, analysts can detect opposition playmakers who pose an unpredictable creative threat, requiring targeted defensive strategies. Alternatively, they can identify players who demonstrate a high tendency for decision-making errors, designating them as optimal targets for coordinated pressing structures.

\section{Conclusions}
\label{section:conclusions}

This paper presented a novel framework for evaluating decision-making in professional football, marking a critical shift from traditional outcome-based statistics to intent-driven spatial analysis. Historically, football analytics has struggled to quantify the quality of a passing decision, often prioritizing safe, lateral completions over tactically ambitious, high-value opportunities. By framing receiver selection as a complex, edge-conditioned graph problem, the proposed Football Pass MPNN successfully bridges the gap between basic pass probability heuristics and advanced geometric deep learning. \\
By introducing a star graph topology centered on the passer, this research demonstrated that opponent pressure and geometrical occlusion can be explicitly and efficiently encoded into node and edge features. Rather than relying on computationally heavy, fully connected networks, the proposed architecture leverages iterative message-passing to contextually evaluate the tactical viability of all teammates in real-time. The empirical results obtained from this approach are consistent with the model's performance, which has been shown to achieve a Top-1 Accuracy of 75.83\% and a Top-3 Accuracy of 97.80\%.\\
Beyond its theoretical and statistical achievements, this framework offers immediate, operational value to the sports industry. The formalisation of intuitive tactical metrics, such as the Best Pass Score, Good Pass Score, and Creativity Ratio, facilitates the translation of abstract predictive probabilities into actionable insights. As demonstrated through its integration within the Data Department of the Royal Belgian Football Association, the MPNN has been shown to allow analysts to rapidly identify suboptimal decision-making, evaluate player risk-reward profiles, and optimise pre-match opposition scouting.\\
Future development in this field might be focused on coach-adaptive fine-tuning, where the model is updated based on a team's recent match history to align with a particular coaching philosophy. This framework could evolve from a general-purpose analytical tool into a personalized decision-support system, more deeply aligned with the intended game model of an individual coaching staff, through conditional probability modelling of recurring tactical patterns and preferred build-up sequences.\\

\section*{Declarations}

\subsubsection*{Acknowledgements and Funding}
G.A.D. and G.R. acknowledge the support provided by the European Union -- NextGenerationEU, in the framework of the iNEST -- Interconnected Nord-Est Innovation Ecosystem (iNEST ECS00000043 – CUP G93C22000610007) project and its CC5 Young Researchers initiative. 
The views and opinions expressed are solely those of the authors and do not necessarily reflect those of the European Union, nor can the European Union be held responsible for them. G.A.D. and G.R. would like to acknowledge INdAM–GNCS.
All expenses related to the acquisition of the data, deployment of the model on the cloud, and all related computational resources were covered by the Royal Belgian Football Association.

\subsubsection*{Conflict of interest/Competing interests}
Not Applicable

\subsubsection*{Ethics approval and consent to participate}
Not Applicable

\subsubsection*{Consent for publication}
Not Applicable

\subsubsection*{Materials availability}
Not Applicable

\subsubsection*{Data availability}
Private data possessed by the Royal Belgian Football Association.

\subsubsection*{Code availability}
Not available for public use.

\subsubsection*{Author contribution}
\textit{Gabriel Masella} was the primary author, responsible for conducting the principal research, developing the codebase, and writing the manuscript.\\
\textit{Giuseppe Alessio D'Inverno} contributed significantly to the project by offering extensive academic support, incorporating specialized graph theory knowledge, validating the final outcomes, and refining the submitted manuscript.\\
\textit{Max Goldsmith} supervised the early stages of the project at the Royal Belgian Football Association, validating initial results and providing foundational guidance on model design.\\
\textit{Gianluigi Rozza} provided supervision and the approval of the overall project, offering also critical resources and mathematical expertise through the MathLAB research group.







\bibliography{sn-bibliography}

\end{document}